\documentclass[letterpaper, 10pt, conference]{IEEEtran}  

\IEEEoverridecommandlockouts                              

\usepackage{graphicx} 
\usepackage{subcaption}
\usepackage{latexsym}
\usepackage{multirow}

\usepackage{hyperref}
\hypersetup{colorlinks}

\usepackage{amsmath}

\usepackage{booktabs}

\usepackage{url}

\usepackage{tikz}
\usepackage{pgfplots}

\usetikzlibrary{calc}
\usetikzlibrary{positioning}
\usepgflibrary{fpu}
\usepgfplotslibrary{fillbetween}

\usetikzlibrary{external}
\pgfplotsset{compat=newest}

\newenvironment{customlegend}[1][]{%
    \begingroup
    \csname pgfplots@init@cleared@structures\endcsname
    \pgfplotsset{#1}%
}{%
    \csname pgfplots@createlegend\endcsname
    \endgroup
}%
\def\addlegendimage{\csname pgfplots@addlegendimage\endcsname}

\title{\LARGE \bf
Pre-trained Models for Sonar Images
}

\author{Matias Valdenegro-Toro$^{1}$
    \and
    Alan Preciado-Grijalva$^{1,2}$
    \and
    Bilal Wehbe$^{1}$
\thanks{This work has been partially supported by the H2020-ICT-2020-2 ICT-47-2020 project DeeperSense (Ref 101016958)
        funded by the European Union’s Horizon 2020 research and innovation programme.
        
        $^{1}$German Research Center for Artificial Intelligence, 28359 Bremen, Germany
        {\tt\small matias.valdenegro@dfki.de}%
        
        $^{2}$Bonn-Rhein-Sieg University of Applied Sciences, 53757 Sankt Augustin, Germany
        {\tt\small alan.preciado@smail.inf.h-brs.de}}%
}

\begin{document}

\maketitle

\begin{abstract}
    Machine learning and neural networks are now ubiquitous in sonar perception, but it lags behind the computer vision field due to the lack of data and pre-trained models specifically for sonar images.
    In this paper we present the Marine Debris Turntable dataset and produce pre-trained neural networks trained on this dataset, meant to fill the gap of missing pre-trained models for sonar images.
    We train Resnet 20, MobileNets, DenseNet121, SqueezeNet, MiniXception, and an Autoencoder, over several input image sizes, from $\mathbf{32 \times 32}$ to $\mathbf{96 \times 96}$, on the Marine Debris turntable dataset.
    We evaluate these models using transfer learning for low-shot classification in the Marine Debris Watertank and another dataset captured using a Gemini 720i sonar. Our results show that in both datasets the pre-trained models produce good features that allow good classification accuracy with low samples (10-30 samples per class). The Gemini dataset validates that the features transfer to other kinds of sonar sensors.
    We expect that the community benefits from the public release of our pre-trained models and the turntable dataset.
\end{abstract}

\section{Introduction}

Machine learning is now ubiquitous in processing sonar images and performing perception of autonomous underwater vehicles. But it is limited by the size and quality of available training data, as capturing datasets with sonar images is more difficult than standard optical images.

In computer vision, pre-trained convolutional neural networks are used as a way to inject prior knowledge into a task \cite{huh2016makes}. These models are usually trained on large datasets such as ImageNet \cite{russakovsky2015imagenet} or OpenImages \cite{kuznetsova2020open}, learning high quality features that are reusable for other tasks \cite{sharif2014cnn} such as object detection, semantic segmentation, etc, through transfer learning.

The application of transfer learning in sonar images is possible \cite{valdenegro2017best} \cite{fuchs2018object}, but it is limited by the availability of data and pre-trained models. Researchers do not generally share the models they train on their datasets, limiting advances in sonar perception.

There is clear need for pre-trained standard architectures with sonar images, which would reduce the sample complexity of classification tasks, serve as strong baselines for future work, and improve performance for other tasks like object detection or semantic segmentation, all with an architecture that is particularly tuned for sonar images, with one channel image data instead of the typical three channels (red, green, and blue). This motivates a tuning of particular architectures, usually by reducing the number of filters and their overall width.

For this purpose, we produce pre-trained models of standard neural network architectures, namely ResNet20 \cite{he2016deep}, MobileNets \cite{howard2017mobilenets}, DenseNet121 \cite{densenet}, SqueezeNet \cite{iandola2016squeezenet}, and MiniXception \cite{arriaga2017real} on a turntable dataset with high variability captured using an ARIS Explorer 3000 Forward-Looking Sonar. We release\footnote{Available at \url{https://github.com/mvaldenegro/pretrained-models-sonar-images}} these models to the community, as a way to kickstart future research into computer vision with sonar images, in a similar way that models pre-trained on the ImageNet dataset have done for color images.

Our contributions are: we produce publicly available pre-trained convolutional neural networks for sonar images, at varying input image resolutions (from $32 \times 32$ to $96 \times 96$). Pre-training happens in a new dataset captured using an ARIS Explorer 3000 FLS, with objects in a rotating platform, allowing the capture of all object views. We call this dataset the Marine Debris Turntable dataset. We demonstrate the potential of our pre-trained models in two image classification tasks, one on the Marine Debris Watertank dataset \footnote{This dataset is available at \url{https://github.com/mvaldenegro/marine-debris-fls-datasets/releases/}}, and another in a binary classification dataset captured using a Gemini 720i Forward-Looking Sonar, which acts as an independent dataset.

We expect that the release of our turntable dataset and the pre-trained models in this dataset can help other researchers advance the field of computer vision on sonar images.

\section{State of the Art}

Transfer learning has been a standard approach to improve performance in computer vision systems \cite{sharif2014cnn} for a long time, but these improvements have not been transferred to the acoustic imaging domain, mostly due to the lack of pre-trained models trained specifically for sonar images.

Valdenegro-Toro \cite{valdenegro2017best} showed that transfer learning in sonar images helps reduce sample complexity, with a slight performance loss if the transfer and feature learning datasets do not share objects in common.

Fuchs et al. \cite{fuchs2018object} uses off the shelf models pre-trained on color images (ImageNet specifically), showing that these also improve performance in sonar image classification, but these models are very big, and require hacks in order to work with a sonar image (transforming from one to three channels), which wastes precious power and computation.

Lou et al. \cite{lou2020automatic} uses GradCam in order to infuse localization capabilities in a ResNet18 trained for color images and to also generalize and improve performance on sonar images as evaluated on the watertank dataset. This can be considered an advanced method of transfer learning as it combines features from optical and sonar data.

\section{Pre-Trained Model Selection}

This section describes the architectures we selected to build pre-trained models:

\textbf{ResNet}. This model uses residual connections in order to improve gradient propagation through the network, and overall it allows much deeper networks. We use the ResNet20 variant, with 20 residual layers, which produce high quality features.

\textbf{MobileNet}. This architecture is designed to be lightweight and allow fast inference in embedded and mobile systems. It uses depthwise separable convolutions in order to reduce computation, which induces a trade-off between task performance (like accuracy) and computational performance. MobileNets are a good choice for resource constrained platforms like Autonomous Underwater Vehicles (AUVs), and the community would benefit from a version of this architecture trained for sonar images.

\textbf{DenseNet}. This architecture makes heavy use of feature reuse, with a set of dense convolutional blocks, each of them being a series of convolutional layers, where each layer receives as inputs all the feature maps from previous convolutional layers in the same block. This allows for very good task performance with a modest increase in the number of parameters, but it increases the amount of memory and computation required for a forward pass. This network is a good choice when very good task performance is required. We use the DenseNet121 variation which is the shallowest of the networks defined in \cite{densenet}.

\textbf{SqueezeNet}. The aim of this model is to reduce the number of parameters while maintaining competitive task performance, and this is achieved by carefully designing the so called Fire module which contains two sub-modules, one that squeezes information through a bottleneck (width lower than the current number of channels) and another that expands the amount of information. SqueezeNet has performance that is similar to AlexNet on ImageNet, but with overall 50 times less parameters, making it a good choice for task performance vs computation trade-off.

\textbf{MiniXception}. This network is a specific design of the Xception network to reduce the amount of computation, in particular for facial emotion recognition. Xception is another design that combines insights from MobileNets by using separable convolutions, into the design of the Inception network, which also performs quite well on the ImageNet competition.

\textbf{Autoencoders}. We train a Convolutional Autoencoder, which can be done without supervision, with the the following encoder architecture Conv2D(32, $3 \times 3$)-MaxPool($2 \times 2$)-Conv2D(16, $3 \times 3$)-MaxPool($2 \times 2$)-Conv2D(8, $3 \times 3$)-MaxPool($2 \times 2$)-Flatten()-FC($c$), and decoder architecture: FC($c n_w n_h$)-Reshape()-Conv2D(32, $3 \times 3$)-UpSample($2 \times 2$)-Conv2D(16, $3 \times 3$)-UpSample($2 \times 2$)-Conv2D(8, $3 \times 3$)-UpSample($2 \times 2$)-Conv2D(1, $3 \times 3$). This architecture is trained using a mean squared error loss, and the Adam optimizer, with the same setup as the other architectures. For feature extraction, only the encoder is used. The feature or code size $c$ can be varied to produce different feature sizes. We vary $c \in [4, 8, 16, 32, 64, 128]$, with each value of $c$ producing a unique autoencoder architecture.

\subsection{Hyper-Parameter Tuning}

As the original architectures we selected are designed for color images and their increased visual complexity, they need to be tuned for the specifics of sonar images, mainly a reduction in the number of parameters (to lessen the chances of overfitting), which can be done to adjusting the width of each layer. Depth of these networks should not be a problem.

\begin{figure*}
    \centering
    \begin{tikzpicture}
        \begin{customlegend}[legend columns = 3,legend style = {column sep=1ex}, legend cell align = left,
            legend entries={Mean, Maximum, Minimum}]
            \addlegendimage{mark=none,red, only marks}
            \addlegendimage{mark=none,green, only marks}
            \addlegendimage{mark=none,blue, only marks}
        \end{customlegend}
    \end{tikzpicture}

    \begin{subfigure}{0.32\textwidth}
        \begin{tikzpicture}
            \begin{axis}[height = 0.18 \textheight, width = 0.9\textwidth, xlabel={Width ($M$)}, ylabel={CV Accuracy (\%)}, ymajorgrids=false, xmajorgrids=false, grid style=dashed, legend pos = north east, legend style={font=\scriptsize}, tick label style={font=\scriptsize}, xtick=data]
                
                \addplot+[mark = none, red] table[x  = width_param, y  = acc_mean, col sep = semicolon] {data/width-tuning-resnet20-results-platform.csv};
                
                \addplot[name path=upper, draw=none] table[x = width_param, y expr = \thisrow{acc_mean} + 0.5 * \thisrow{acc_std}, col sep = semicolon] {data/width-tuning-resnet20-results-platform.csv};
                \addplot[name path=lower, draw=none] table[x = width_param, y expr = \thisrow{acc_mean} - 0.5 * \thisrow{acc_std}, col sep = semicolon] {data/width-tuning-resnet20-results-platform.csv};
                \addplot [fill=red!30] fill between[of=upper and lower];
                
                \addplot+[mark = none, green] table[x  = width_param, y  = acc_max, col sep = semicolon] {data/width-tuning-resnet20-results-platform.csv};
                
                \addplot+[mark = none, blue] table[x  = width_param, y  = acc_min, col sep = semicolon] {data/width-tuning-resnet20-results-platform.csv};
                
            \end{axis}		
        \end{tikzpicture}
        \caption{ResNet20}
    \end{subfigure}
        \begin{subfigure}{0.32\textwidth}
            \begin{tikzpicture}
                \begin{axis}[height = 0.18 \textheight, width = 0.9\textwidth, xlabel={Width ($M$)}, ylabel={CV Accuracy (\%)}, ymajorgrids=false, xmajorgrids=false, grid style=dashed, legend pos = north east, legend style={font=\scriptsize}, tick label style={font=\scriptsize}, xtick=data]
                    
                    \addplot+[mark = none, red] table[x  = width_param, y  = acc_mean, col sep = semicolon] {data/width-tuning-mobilenet-results-platform.csv};
                    
                    \addplot[name path=upper, draw=none] table[x = width_param, y expr = \thisrow{acc_mean} + 0.5 * \thisrow{acc_std}, col sep = semicolon] {data/width-tuning-mobilenet-results-platform.csv};
                    \addplot[name path=lower, draw=none] table[x = width_param, y expr = \thisrow{acc_mean} - 0.5 * \thisrow{acc_std}, col sep = semicolon] {data/width-tuning-mobilenet-results-platform.csv};
                    \addplot [fill=red!30] fill between[of=upper and lower];
                    
                    \addplot+[mark = none, green] table[x  = width_param, y  = acc_max, col sep = semicolon] {data/width-tuning-mobilenet-results-platform.csv};
                    
                    \addplot+[mark = none, blue] table[x  = width_param, y  = acc_min, col sep = semicolon] {data/width-tuning-mobilenet-results-platform.csv};
                    
                \end{axis}		
            \end{tikzpicture}
            \caption{MobileNets}
        \end{subfigure}
        \begin{subfigure}{0.32\textwidth}
            \begin{tikzpicture}
                \begin{axis}[height = 0.18 \textheight, width = 0.9\textwidth, xlabel={Width ($M$)}, ylabel={CV Accuracy (\%)}, ymajorgrids=false, xmajorgrids=false, grid style=dashed, legend pos = north east, legend style={font=\scriptsize}, tick label style={font=\scriptsize}, xtick=data]
                    
                    \addplot+[mark = none, red] table[x  = width_param, y  = acc_mean, col sep = semicolon] {data/width-tuning-squeezenet-results-platform.csv};
                    
                    \addplot[name path=upper, draw=none] table[x = width_param, y expr = \thisrow{acc_mean} + 0.5 * \thisrow{acc_std}, col sep = semicolon] {data/width-tuning-squeezenet-results-platform.csv};
                    \addplot[name path=lower, draw=none] table[x = width_param, y expr = \thisrow{acc_mean} - 0.5 * \thisrow{acc_std}, col sep = semicolon] {data/width-tuning-squeezenet-results-platform.csv};
                    \addplot [fill=red!30] fill between[of=upper and lower];
                    
                    \addplot+[mark = none, green] table[x  = width_param, y  = acc_max, col sep = semicolon] {data/width-tuning-squeezenet-results-platform.csv};
                    
                    \addplot+[mark = none, blue] table[x  = width_param, y  = acc_min, col sep = semicolon] {data/width-tuning-squeezenet-results-platform.csv};
                    
                \end{axis}		
            \end{tikzpicture}
            \caption{SqueezeNet}
        \end{subfigure}
    
        \begin{subfigure}{0.32\textwidth}
            \begin{tikzpicture}
                \begin{axis}[height = 0.18 \textheight, width = 0.9\textwidth, xlabel={Width ($M$)}, ylabel={CV Accuracy (\%)}, ymajorgrids=false, xmajorgrids=false, grid style=dashed, legend pos = north east, legend style={font=\scriptsize}, tick label style={font=\scriptsize}, xtick=data]
                    
                    \addplot+[mark = none, red] table[x  = width_param, y  = acc_mean, col sep = semicolon] {data/width-tuning-densenet121-results-platform.csv};
                    
                    \addplot[name path=upper, draw=none] table[x = width_param, y expr = \thisrow{acc_mean} + 0.5 * \thisrow{acc_std}, col sep = semicolon] {data/width-tuning-densenet121-results-platform.csv};
                    \addplot[name path=lower, draw=none] table[x = width_param, y expr = \thisrow{acc_mean} - 0.5 * \thisrow{acc_std}, col sep = semicolon] {data/width-tuning-densenet121-results-platform.csv};
                    \addplot [fill=red!30] fill between[of=upper and lower];
                    
                    \addplot+[mark = none, green] table[x  = width_param, y  = acc_max, col sep = semicolon] {data/width-tuning-densenet121-results-platform.csv};
                    
                    \addplot+[mark = none, blue] table[x  = width_param, y  = acc_min, col sep = semicolon] {data/width-tuning-densenet121-results-platform.csv};
                    
                \end{axis}		
            \end{tikzpicture}
            \caption{DenseNet121}
        \end{subfigure}    
        \begin{subfigure}{0.32\textwidth}
            \begin{tikzpicture}
                \begin{axis}[height = 0.18 \textheight, width = 0.9\textwidth, xlabel={Width ($M$)}, ylabel={CV Accuracy (\%)}, ymajorgrids=false, xmajorgrids=false, grid style=dashed, legend pos = north east, legend style={font=\scriptsize}, tick label style={font=\scriptsize}, xtick=data]
                    
                    \addplot+[mark = none, red] table[x  = width_param, y  = acc_mean, col sep = semicolon] {data/width-tuning-minixception-results-platform.csv};
                    
                    \addplot[name path=upper, draw=none] table[x = width_param, y expr = \thisrow{acc_mean} + 0.5 * \thisrow{acc_std}, col sep = semicolon] {data/width-tuning-minixception-results-platform.csv};
                    \addplot[name path=lower, draw=none] table[x = width_param, y expr = \thisrow{acc_mean} - 0.5 * \thisrow{acc_std}, col sep = semicolon] {data/width-tuning-minixception-results-platform.csv};
                    \addplot [fill=red!30] fill between[of=upper and lower];
                    
                    \addplot+[mark = none, green] table[x  = width_param, y  = acc_max, col sep = semicolon] {data/width-tuning-minixception-results-platform.csv};
                    
                    \addplot+[mark = none, blue] table[x  = width_param, y  = acc_min, col sep = semicolon] {data/width-tuning-minixception-results-platform.csv};
                    
                \end{axis}		
            \end{tikzpicture}
            \caption{MiniXception}
        \end{subfigure}
    \caption{Results from 5-Fold Cross validation on the Turntable training set, showing the effect of the width on accuracy for each selected architecture. Shaded areas represents one-$\sigma$ confidence intervals.}
    \label{cv_turntable_architectures}
\end{figure*}
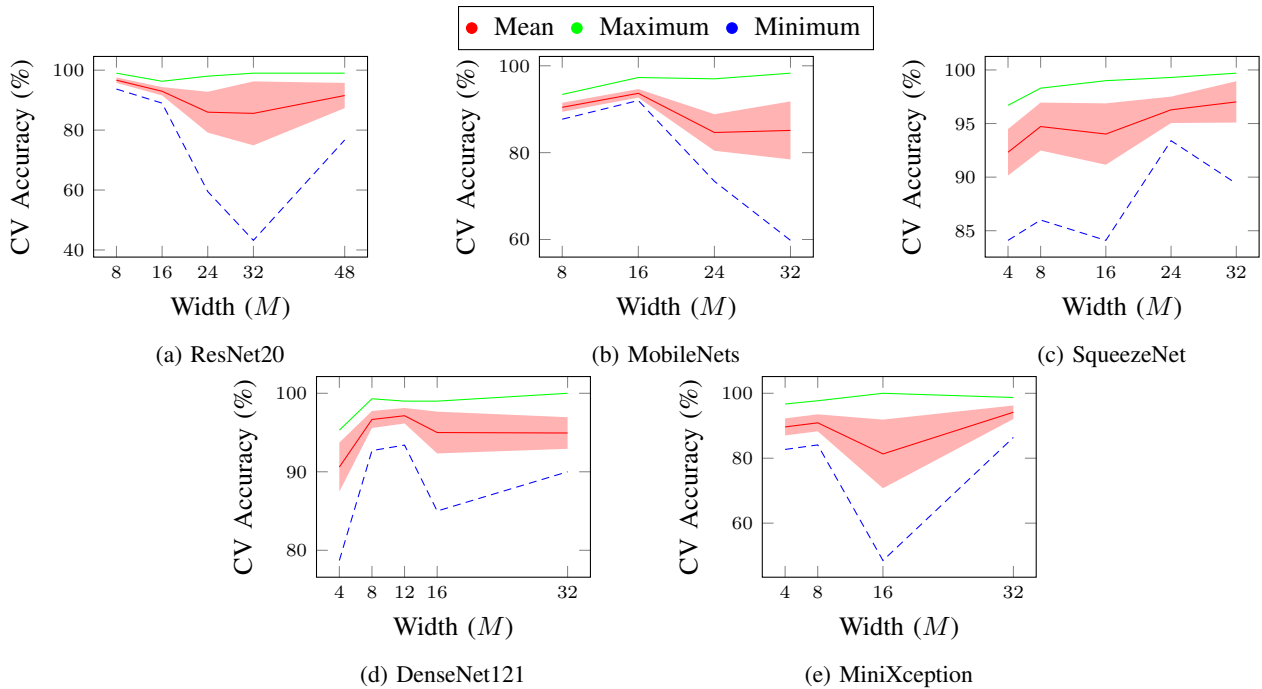

Each architecture is slightly modified. We choose the shallowest variations of each architecture, and we reduce the width (by changing the number of filters or neurons) on each architecture, tuning them to produce maximum task performance, which usually happens with less than 128 filters (instead of up to 1024 filters in the original architecture). For this purpose we perform a grid search to decide the width of each architecture, on the Marine Debris Turntable training set at resolution $96 \times 96$, using K-Fold Cross Validation with $k = 5$. We evaluate widths varying in the range $8-64$, depending on the architecture. These results are shown in Figure \ref{cv_turntable_architectures}.

We selected the width of each architecture to maximize mean test accuracy over folds in K-Fold cross-validation. The selected widths are displayed in Table \ref{selected_widths}, as well as how the width $w$ parameterize each architecture.

\subsection{Input Image Variations}

For each architecture, we train a set of models varying the input image size, including sizes $s \times s$ for $s \in [32, 48, 64, 80, 96]$, as the original input size typically used in the ImageNet dataset ($224 \times 224$) is too big for a sonar image, depending on frequency. This allows to learn features at multiple scales, and to evaluate the optimal image size. We will provide all models trained as the end user can select the best scale depending on their needs. Note that smaller scales also require less computation.

\begin{table}
    \centering
    \begin{tabular}{llp{4cm}}
        \toprule
        Architecture	&	Selected Width $w$	& Use of $w$ in Architecture\\
        \midrule
        Resnet20		& 32				& $w$ is set as starting number of filters, duplicating itself after each residual stack ($w = 2 w$).\\
        MobileNet		& 32				& Base filters are set to $w$, increasing first by a factor of four ($w = 4w$), then duplicating themselves.\\
        DenseNet121		& 16				& $w$ is used as number of filters for each convolutional layer inside each dense block.\\
        SqueezeNet		& 32				& Base squeeze filters are set to $w$ , and expansion filters are set to $2w$, duplicating itself after each fire module stack.\\
        MiniXception	& 16				& Stem widths are set to $[\frac{w}{2}, w]$, and block widths to $[w, 2w, 4w, 8w]$.\\
        \bottomrule
    \end{tabular}
    \caption{Selected width hyper-parameter for each pre-trained architecture, based on results from Figure \ref{cv_turntable_architectures}.}
    \label{selected_widths}
\end{table}

\subsection{Computational Performance}

We believe that computational performance is important for the future users of our pre-trained models, as we expect that they will be used for real-time applications such as AUVs, which requires fast processing on low power embedded platforms. In this section we provide information on the computational requirements of each model.

\begin{figure*}
    \centering
    \begin{tikzpicture}
        \begin{customlegend}[legend columns = 5,legend style = {column sep=1ex}, legend cell align = left,
            legend entries={$32 \times 32$, $48 \times 48$, $64 \times 64$, $80 \times 80$, $96 \times 96$}]
            \addlegendimage{mark=none,red, only marks}
            \addlegendimage{mark=none,green, only marks}
            \addlegendimage{mark=none,blue, only marks}
            \addlegendimage{mark=none,black, only marks}
            \addlegendimage{mark=none,purple, only marks}
        \end{customlegend}
    \end{tikzpicture}        

    \begin{subfigure}{0.30\textwidth}
        \begin{tikzpicture}
            \begin{axis}[height = 0.16 \textheight, width = 0.9\textwidth, xlabel={Width ($M$)}, ylabel={FLOPs}, ymajorgrids=false, xmajorgrids=false, grid style=dashed, legend pos = north east, legend style={font=\scriptsize}, tick label style={font=\scriptsize}, xtick=data]
                
                \addplot+[mark = none, red] table[x  = width_param, y  = flops, col sep = semicolon] {data/width-flops-resnet20-imagesize-32x32.csv};
                                
                \addplot+[mark = none, green] table[x  = width_param, y  = flops, col sep = semicolon] {data/width-flops-resnet20-imagesize-48x48.csv};
                
                \addplot+[mark = none, blue] table[x  = width_param, y  = flops, col sep = semicolon] {data/width-flops-resnet20-imagesize-64x64.csv};
                
                \addplot+[mark = none, black] table[x  = width_param, y  = flops, col sep = semicolon] {data/width-flops-resnet20-imagesize-80x80.csv};
                
                \addplot+[mark = none, purple] table[x  = width_param, y  = flops, col sep = semicolon] {data/width-flops-resnet20-imagesize-96x96.csv};
            \end{axis}		
        \end{tikzpicture}
        \caption{ResNet20}
    \end{subfigure}
    \begin{subfigure}{0.30\textwidth}
        \begin{tikzpicture}
            \begin{axis}[height = 0.16 \textheight, width = 0.9\textwidth, xlabel={Width ($M$)}, ylabel={FLOPs}, ymajorgrids=false, xmajorgrids=false, grid style=dashed, legend pos = north east, legend style={font=\scriptsize}, tick label style={font=\scriptsize}, xtick=data]
                
                \addplot+[mark = none, red] table[x  = width_param, y  = flops, col sep = semicolon] {data/width-flops-densenet121-imagesize-32x32.csv};
                
                \addplot+[mark = none, green] table[x  = width_param, y  = flops, col sep = semicolon] {data/width-flops-densenet121-imagesize-48x48.csv};
                
                \addplot+[mark = none, blue] table[x  = width_param, y  = flops, col sep = semicolon] {data/width-flops-densenet121-imagesize-64x64.csv};
                
                \addplot+[mark = none, black] table[x  = width_param, y  = flops, col sep = semicolon] {data/width-flops-densenet121-imagesize-80x80.csv};
                
                \addplot+[mark = none, purple] table[x  = width_param, y  = flops, col sep = semicolon] {data/width-flops-densenet121-imagesize-96x96.csv};
            \end{axis}		
        \end{tikzpicture}
        \caption{DenseNet121}
    \end{subfigure}
    \begin{subfigure}{0.30\textwidth}
        \begin{tikzpicture}
            \begin{axis}[height = 0.16 \textheight, width = 0.9\textwidth, xlabel={Width ($M$)}, ylabel={FLOPs}, ymajorgrids=false, xmajorgrids=false, grid style=dashed, legend pos = north east, legend style={font=\scriptsize}, tick label style={font=\scriptsize}, xtick=data]
                
                \addplot+[mark = none, red] table[x  = width_param, y  = flops, col sep = semicolon] {data/width-flops-squeezenet-imagesize-32x32.csv};
                
                \addplot+[mark = none, green] table[x  = width_param, y  = flops, col sep = semicolon] {data/width-flops-squeezenet-imagesize-48x48.csv};
                
                \addplot+[mark = none, blue] table[x  = width_param, y  = flops, col sep = semicolon] {data/width-flops-squeezenet-imagesize-64x64.csv};
                
                \addplot+[mark = none, black] table[x  = width_param, y  = flops, col sep = semicolon] {data/width-flops-squeezenet-imagesize-80x80.csv};
                
                \addplot+[mark = none, purple] table[x  = width_param, y  = flops, col sep = semicolon] {data/width-flops-squeezenet-imagesize-96x96.csv};
            \end{axis}		
        \end{tikzpicture}
        \caption{SqueezeNet}
    \end{subfigure}

    \begin{subfigure}{0.30\textwidth}
        \begin{tikzpicture}
            \begin{axis}[height = 0.16 \textheight, width = 0.9\textwidth, xlabel={Width ($M$)}, ylabel={FLOPs}, ymajorgrids=false, xmajorgrids=false, grid style=dashed, legend pos = north east, legend style={font=\scriptsize}, tick label style={font=\scriptsize}, xtick=data]
                
                \addplot+[mark = none, red] table[x  = width_param, y  = flops, col sep = semicolon] {data/width-flops-mobilenet-imagesize-32x32.csv};
                
                \addplot+[mark = none, green] table[x  = width_param, y  = flops, col sep = semicolon] {data/width-flops-mobilenet-imagesize-48x48.csv};
                
                \addplot+[mark = none, blue] table[x  = width_param, y  = flops, col sep = semicolon] {data/width-flops-mobilenet-imagesize-64x64.csv};
                
                \addplot+[mark = none, black] table[x  = width_param, y  = flops, col sep = semicolon] {data/width-flops-mobilenet-imagesize-80x80.csv};
                
                \addplot+[mark = none, purple] table[x  = width_param, y  = flops, col sep = semicolon] {data/width-flops-mobilenet-imagesize-96x96.csv};
            \end{axis}		
        \end{tikzpicture}
        \caption{MobileNets}
    \end{subfigure}
    \begin{subfigure}{0.30\textwidth}
        \begin{tikzpicture}
            \begin{axis}[height = 0.16 \textheight, width = 0.9\textwidth, xlabel={Width ($M$)}, ylabel={FLOPs}, ymajorgrids=false, xmajorgrids=false, grid style=dashed, legend pos = north east, legend style={font=\scriptsize}, tick label style={font=\scriptsize}, xtick=data]
                
                \addplot+[mark = none, red] table[x  = width_param, y  = flops, col sep = semicolon] {data/width-flops-minixception-imagesize-32x32.csv};
                
                \addplot+[mark = none, green] table[x  = width_param, y  = flops, col sep = semicolon] {data/width-flops-minixception-imagesize-48x48.csv};
                
                \addplot+[mark = none, blue] table[x  = width_param, y  = flops, col sep = semicolon] {data/width-flops-minixception-imagesize-64x64.csv};
                
                \addplot+[mark = none, black] table[x  = width_param, y  = flops, col sep = semicolon] {data/width-flops-minixception-imagesize-80x80.csv};
                
                \addplot+[mark = none, purple] table[x  = width_param, y  = flops, col sep = semicolon] {data/width-flops-minixception-imagesize-96x96.csv};
            \end{axis}		
        \end{tikzpicture}
        \caption{MiniXception}
    \end{subfigure}
    \caption{Computational performance measured by FLOPs as width and image size is varied.}
    \label{pretrained_flops _performance}
\end{figure*}
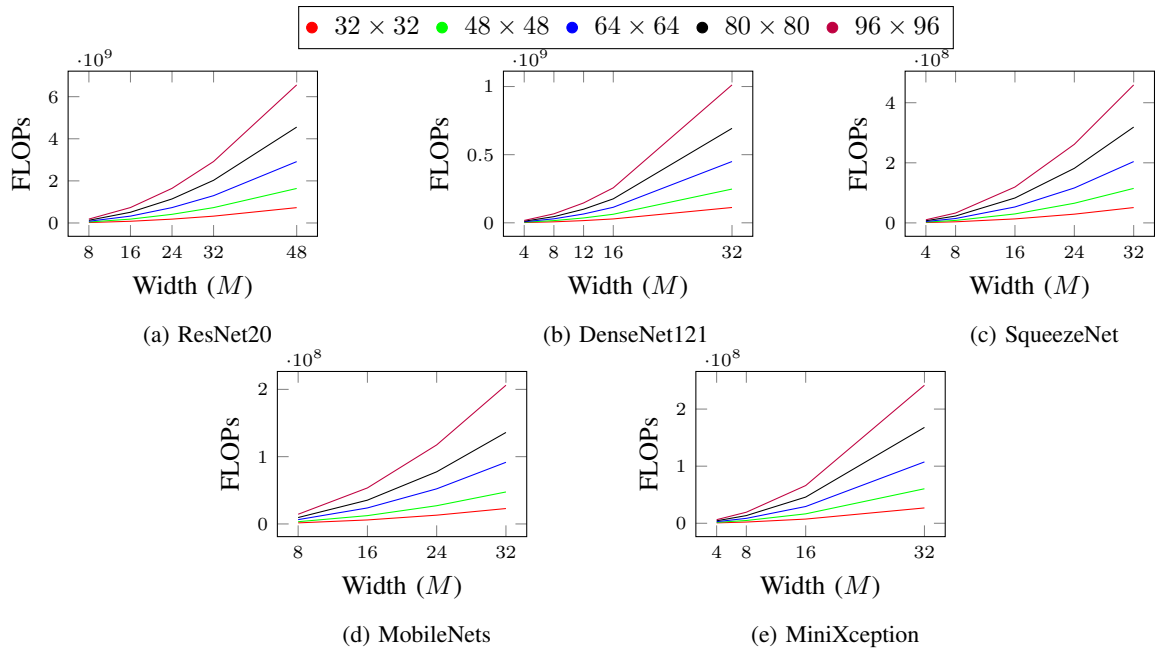

We compute the number of FLOPs\footnote{Floating point operations like additions and multiplications.} required for a single forward pass for each model, as the input image size and widths are varied. These results are shown in Figure \ref{pretrained_flops _performance}.

ResNet20 and DenseNet121 are the most computationally expensive models across all architectures, while MobilNets and MiniXception are the computationally lightest.

\section{Experimental Evaluation}

We train the previously mentioned models (ResNet, SqueezeNet, DenseNet, MobileNet, and MiniXception) on the Marine Debris Turntable Dataset. 

\subsection{Datasets}

\begin{figure*}[t]
    \centering
    \begin{subfigure}{0.15\textwidth}
        \includegraphics[width=\linewidth]{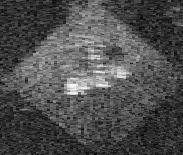}
        \caption{Glass Bottle 1}
    \end{subfigure}
    \begin{subfigure}{0.15\textwidth}
        \includegraphics[width=\linewidth]{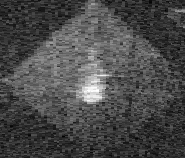}
        \caption{Can}
    \end{subfigure}
    \begin{subfigure}{0.15\textwidth}
        \includegraphics[width=\linewidth]{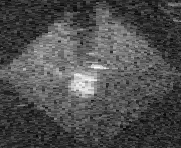}
        \caption{Drink Carton}
    \end{subfigure}
    \begin{subfigure}{0.15\textwidth}
        \includegraphics[width=\linewidth]{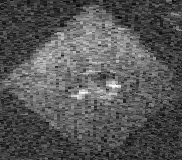}
        \caption{Drink Sachet}
    \end{subfigure}
    \begin{subfigure}{0.15\textwidth}
        \includegraphics[width=\linewidth]{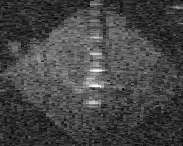}
        \caption{Glass Bottle 2}
    \end{subfigure}
    \begin{subfigure}{0.15\textwidth}
        \includegraphics[width=\linewidth]{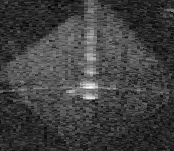}
        \caption{Glass Jar}
    \end{subfigure}
    \begin{subfigure}{0.15\textwidth}
        \includegraphics[width=\linewidth]{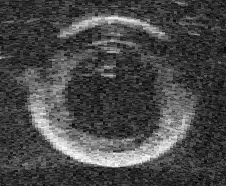}
        \caption{Large Tire}
    \end{subfigure}
    \begin{subfigure}{0.15\textwidth}
        \includegraphics[width=\linewidth]{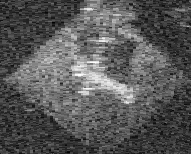}
        \caption{Metal Bottle}
    \end{subfigure}
    \begin{subfigure}{0.15\textwidth}
        \includegraphics[width=\linewidth]{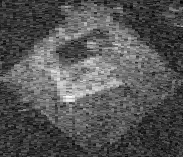}
        \caption{Metal Box}
    \end{subfigure}
    \begin{subfigure}{0.15\textwidth}
        \includegraphics[width=\linewidth]{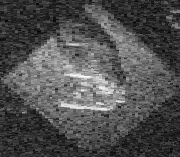}
        \caption{Plastic Bidon}
    \end{subfigure}
    \begin{subfigure}{0.15\textwidth}
        \includegraphics[width=\linewidth]{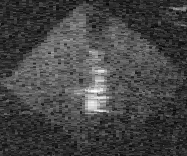}
        \caption{Plastic Bottle}
    \end{subfigure}
    \begin{subfigure}{0.15\textwidth}
        \includegraphics[width=\linewidth]{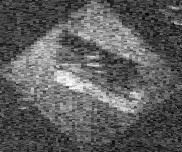}
        \caption{Plastic Pipe}
    \end{subfigure}
    \begin{subfigure}{0.15\textwidth}
        \includegraphics[width=\linewidth]{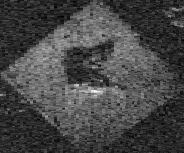}
        \caption{Plastic Propeller}
    \end{subfigure}
    \begin{subfigure}{0.15\textwidth}
        \includegraphics[width=\linewidth]{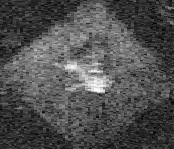}
        \caption{Glass Bottle 3}
    \end{subfigure}
    \begin{subfigure}{0.15\textwidth}
        \includegraphics[width=\linewidth]{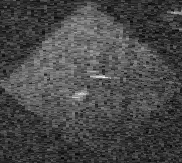}
        \caption{Rotating Platform}
    \end{subfigure}
    \begin{subfigure}{0.15\textwidth}
        \includegraphics[width=\linewidth]{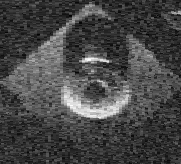}
        \caption{Small Tire}
    \end{subfigure}
    \begin{subfigure}{0.15\textwidth}
        \includegraphics[width=\linewidth]{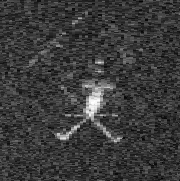}
        \caption{Valve}
    \end{subfigure}
    \begin{subfigure}{0.15\textwidth}
        \includegraphics[width=\linewidth]{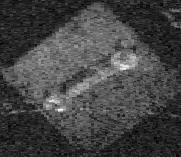}
        \caption{Metal Wrench}
    \end{subfigure}
    \caption{Samples from the Marine Debris Turntable dataset, captured using a ARIS Explorer 3000 FLS, and containing 18 kinds of objects of varying materials and shapes, which are grouped into 12 semantic classes.}
    \label{ds_sample}
\end{figure*}

\begin{figure*}[t]
    \includegraphics[width=0.19\textwidth]{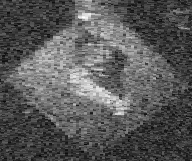}
    \includegraphics[width=0.19\textwidth]{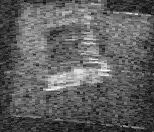}
    \includegraphics[width=0.19\textwidth]{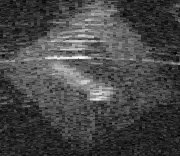}
    \includegraphics[width=0.19\textwidth]{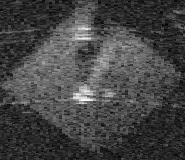}
    \includegraphics[width=0.19\textwidth]{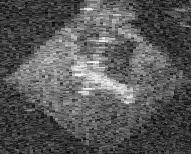}
    \caption{Sample images from the Turntable dataset, showing different views of the metal bottle class, and how it visually changes as the turntable rotates.}
    \label{turntable_metalbottle_samples}
\end{figure*}

We use the following datasets for training and evaluation:

\textbf{Marine Debris Turntable}. This dataset contains 2471 images in total, over 18 different objects that are grouped into 12 classes. The dataset was captured using an ARIS Explorer 3000 forward-looking sonar, at the highest frequency setting. Each object is placed in an underwater turntable and rotated from the minimum to the maximum angle, rotated along the Z axis. This produces a comprehensive set of object views, which is ideal to learn features from, as it contains many properties typical to sonar images that only happen on specific viewpoints, such as reflection artifacts, ambiguous poses, and sensor noise.
A sample of these object classes is shown in Figure \ref{ds_sample}, while samples images of the turntable rotating on the metal box class is shown in Figure \ref{turntable_metalbottle_samples}. The classes in this dataset are: bottle, can, carton, box, bidon, pipe, platform, propeller, sachet, tire, valve, and wrench.

The Turntable dataset has several configurations, and in this work we only use the platform image crop, and 12 classes. The dataset is split into training set with 1505 images, and a test set of 645 images. Note that the original samples of this dataset are stored in the same order as they are captured, which is induced by the turntable rotation. A train-test split is performed across this order for each object as if it was a time series, in order to prevent leakage across the turntable rotation, producing an independent test set.

This dataset will be used for training the architectures and produce pre-trained models for sonar images, as it is the most difficult dataset we have.

\textbf{Marine Debris Watertank}.
This dataset contains 2627 FLS images captured with an ARIS Explorer 3000 forward-looking sonar at 3.0 MHz, split three way into training, validation, and test set \cite{valdenegro2019deep}. In this work we only use the training set containing 1838 images, and the test set of 395 images. The dataset contains 11 classes including the background class. This dataset is used for evaluation of transfer learning using the pre-trained models that we propose in this paper. There is some intersection between the object classes in the Turntable and Watertank dataset (around 50\% of objects), so it is not a completely independent dataset for evaluation.

\begin{figure}[t]
    \centering
    \begin{subfigure}{0.49\textwidth}
        \centering
        \includegraphics[height=0.10\textheight]{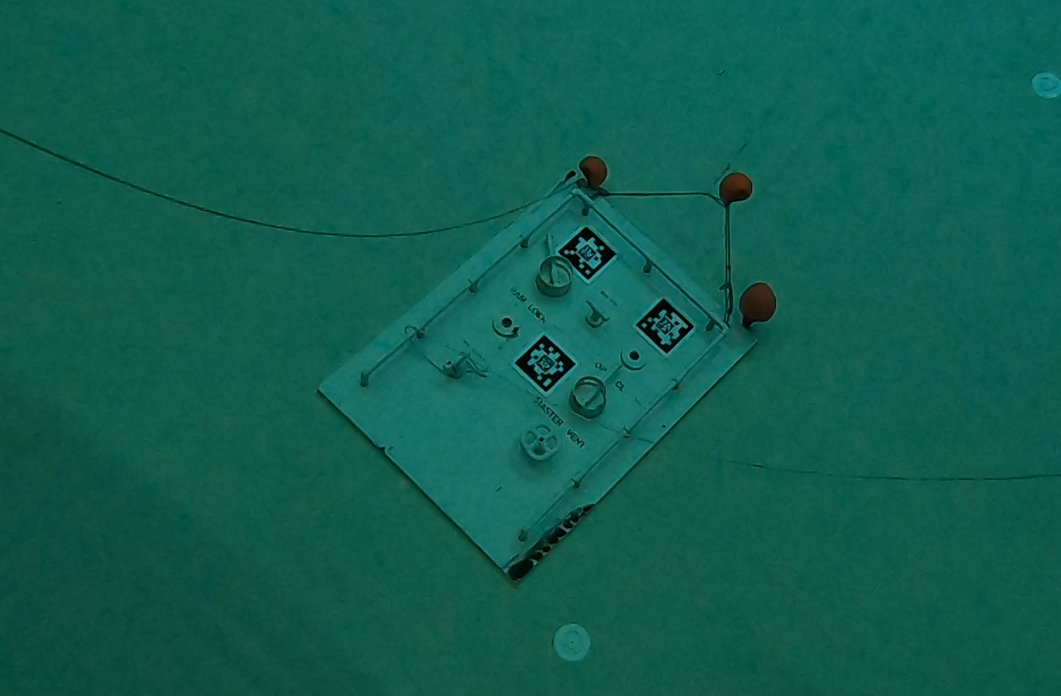}
        \includegraphics[height=0.10\textheight]{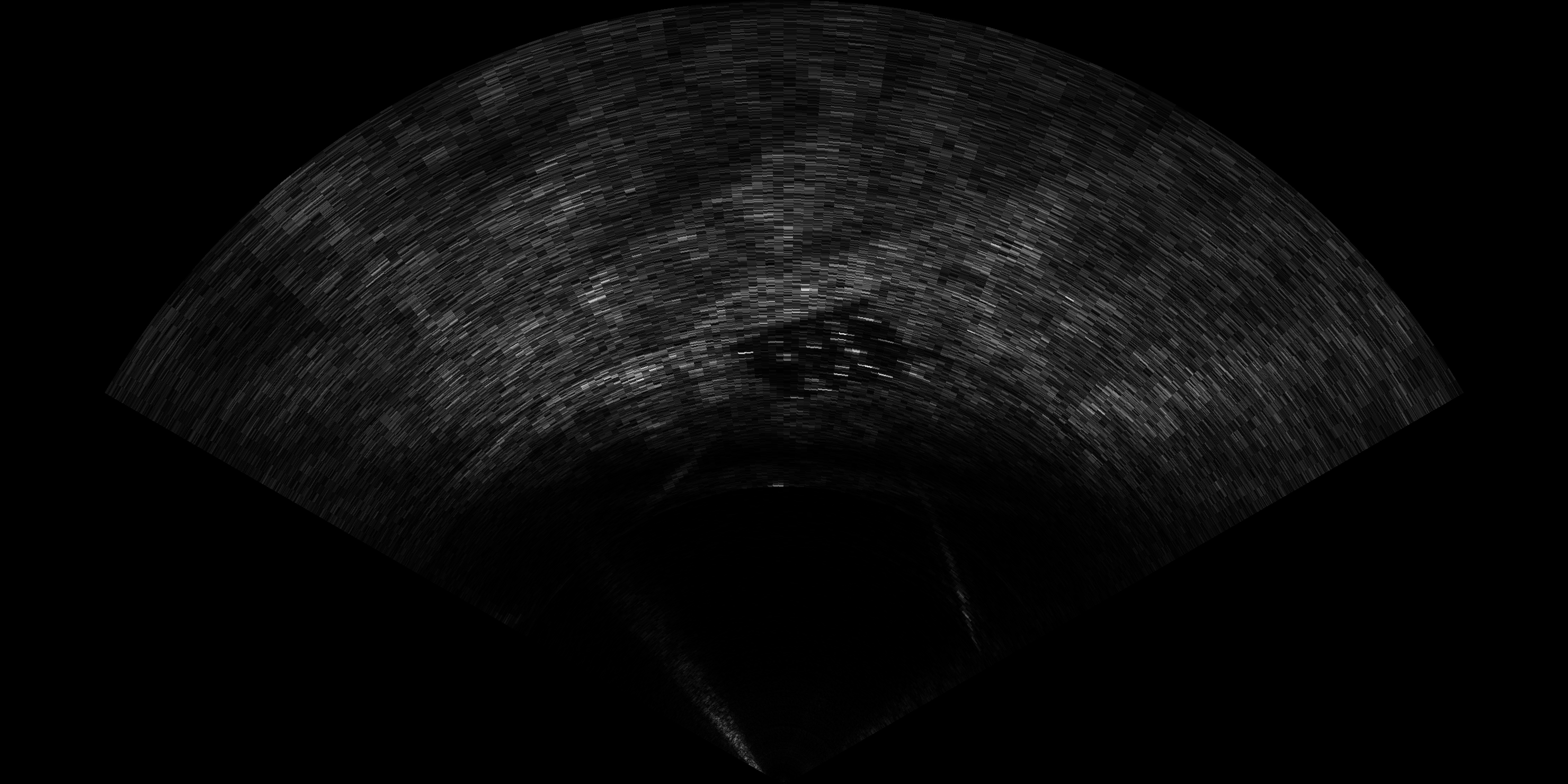}
        \caption{Panel}
    \end{subfigure}
    
    \begin{subfigure}{0.49\textwidth}
        \centering
        \includegraphics[height=0.094\textheight]{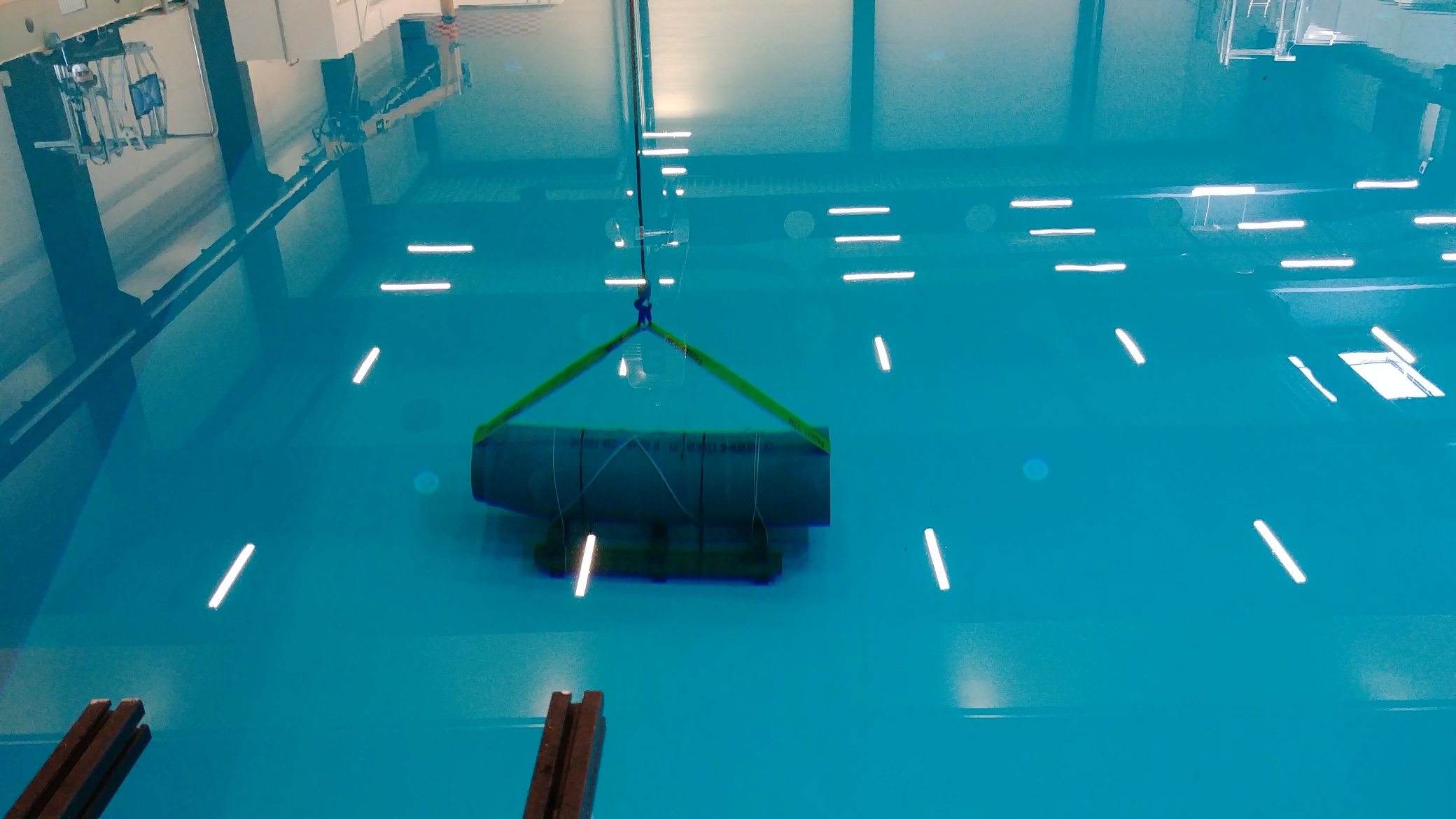}
        \includegraphics[height=0.094\textheight]{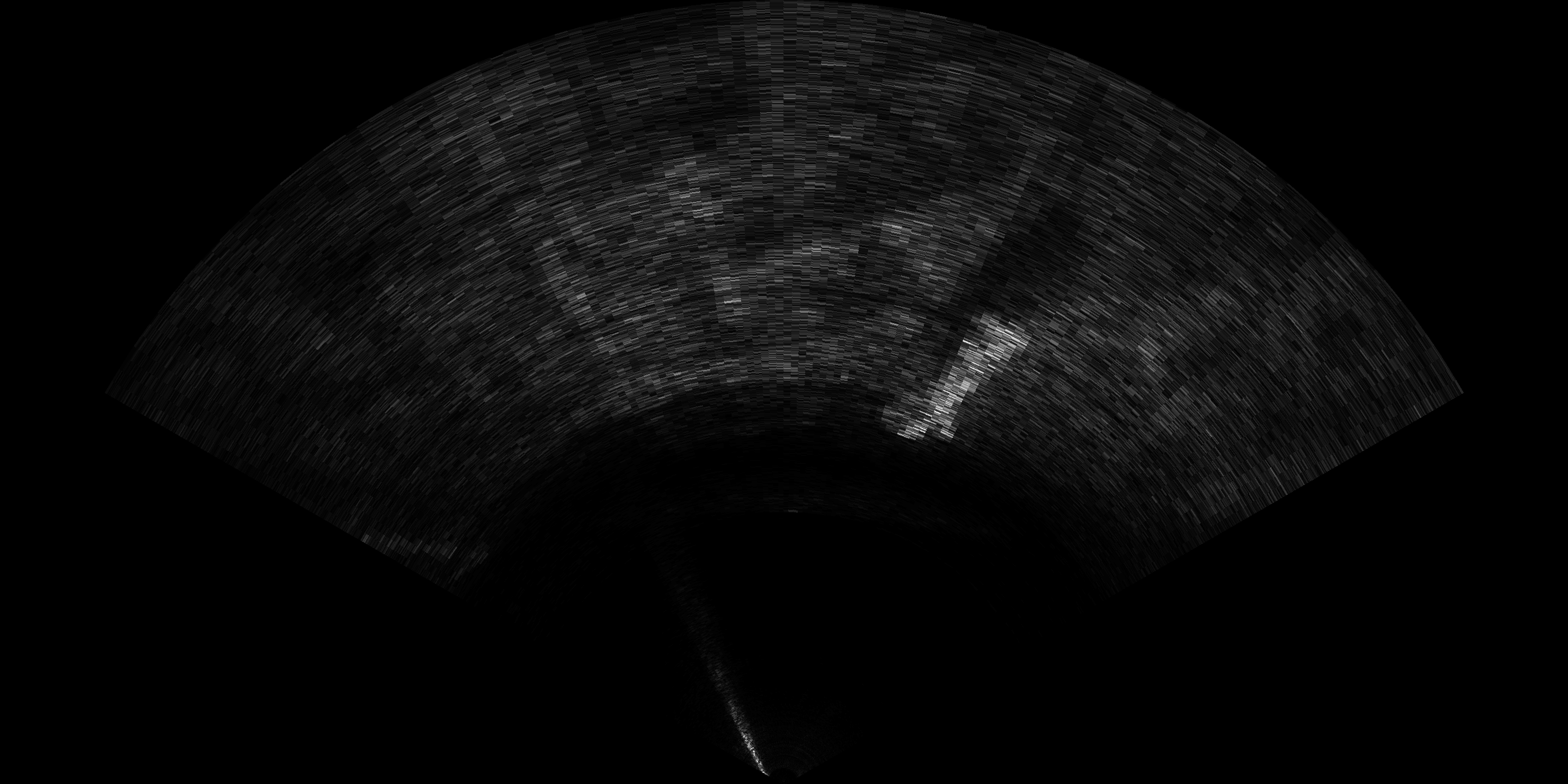}
        \caption{Pipe}
    \end{subfigure}
    \caption{Sample of optical and sonar images from the Gemini Panel-Pipe dataset.}
    \label{gemini_sample}
\end{figure}

\textbf{Gemini 720i Panel-Pipe}. This dataset contains 600 images captured using a Tritech Gemini 720i FLS. We captured two objects, a concrete pipe, and a hard plastic panel, leading to only two semantic classes. The training set is composed of 450 images, and the test set of 150 images. The dataset is mostly balanced, with 290 images for panel and 310 images for concrete pipe. One sample for each of these objects in optical and sonar modalities is shown in Figure \ref{gemini_sample}. This dataset is used as a independent evaluation of transfer learning of our pre-trained models.

Both the Watertank and Turntable datasets are publicly available at \url{https://github.com/mvaldenegro/marine-debris-fls-datasets/releases/}

\subsection{Model Training}

We train each architecture using the Adam optimizer with a learning rate $\alpha = 0.001$, for 200 epochs. We use a single set of data augmentations for all architectures, corresponding to random shift $s_w, s_h$ along the width and height dimensions, sampled from a uniform distribution $s_w \sim \mathrm{U}(0, 0.1 w)$ and $s_h \sim \mathrm{U}(0, 0.1 h)$, and random up-down and left-right flips with 50\% probability each.

Normalization of this dataset proved to be a bit difficult, as the standard normalization by dividing pixel values by $255$ did not work, and we resorted to subtracting the pixel mean computed on the training set, which allowed for all models to train correctly. We used the mean value $\mu_p = 84.5$. Note that this mean varies slightly depending on the input image size. Normalization can be applied using Eq \ref{normalization}.

\begin{equation}
    x_{\text{norm}} = x - \mu_p
    \label{normalization}
\end{equation}

Assuming that the input image $x$ contain pixels in the range $[0, 255]$. The training set and any new images are normalized in the same way.

Results of training all architectures on the Marine Debris Turntable dataset is shown in Table \ref{results_input_image}. The best performing models overall are DenseNet and SqueezeNet. Highest accuracies are obtained with the largest input image sizes, while the smallest image size ($32 \times 32$) is quite difficult to classify for ResNet and SqueezeNet, but the overall trend is that performance increases with the input image size.

\begin{table}[t]
    \centering
    \begin{tabular}{llllll}
        \toprule
        & \multicolumn{5}{c}{Test Accuracy}\\
        Model 			& \rotatebox{45}{$32\times 32$} &  \rotatebox{45}{$48 \times 48$} &  \rotatebox{45}{$64 \times 64$} &  \rotatebox{45}{$80 \times 80$} &  \rotatebox{45}{$96 \times 96$}\\
        \midrule
        ResNet20		& 73.2\% 		& 96.8\% 		 & \textbf{99.1}\%   & 90.9\%			& 93.6\%\\
        MobileNet		& 95.7\% 		& 97.4\% 		 & 96.9\% 		  & 94.4\%			& 96.7\%\\
        DenseNet121		& \textbf{97.8}\% & \textbf{99.4}\% & 98.9\% 		  & \textbf{99.1}\%			& \textbf{99.4}\%\\
        SqueezeNet		& 92.6\% 		& 98.8\% 		 & \textbf{99.4}\%   & 90.7\%			& \textbf{99.7}\%\\
        MiniXception	& 96.9\%		& 81.6\%		 & 61.9\%		  & 96.0\%			& 93.8\%\\
        \midrule
        Linear SVM		& 93.3\%		& 96.3\%		 & 97.5\%		& 97.5\%	& 97.4\%\\
        \toprule
        Image mean $\mu_p$		& $84.54$		& $84.53$		 & $84.51$		  & $84.51$			& $84.51$\\
        \toprule
    \end{tabular}
    \caption{Classification results on the test set of the Marine Debris Turntable dataset, as the input image size is varied. The bottom row contains the detailed image mean used to train each of the models at the given input image size.}
    \label{results_input_image}
\end{table}

\subsection{Transfer Learning on Watertank Dataset}

For each architecture, we select the $96 \times 96$ input image variation, and we perform transfer learning, targeting the Marine Debris Watertank dataset. 

As transfer learning generally improves sample complexity (higher accuracy with less training samples), we sub-sample the training set to contain $\text{spc} \in [1, 5, 10, 20, 30, 40, 50]$ samples for each class, which simulates a low-shot setting. We extract features in the sub-sampled training set, using a selection of layers close to the output in each architecture (usually the Flatten or last ReLU/Batch Normalization layer), and use them to train a support vector machine with $C = 1.0$ and a linear kernel. We perform 10 runs of each experiment (value of $\text{spc}$) and present the mean and standard deviation of accuracy.

The purpose of using a linear SVM is to evaluate the quality of the features, without going into difficulties with the optimization process or tuning hyper-parameters of new fully connected layers or kernels for SVMs. It is a simple test for linear separability.

To quantify the sample complexity of each model, we plot test accuracy vs the number of samples in the training set, producing a curve we call the sample complexity curve, and the area under this curve defines a sample complexity metric. We call this metric area under the sample complexity curve (AUSCC), but note that this is not the same AUC as in the ROC curve. This allows to select the model that best performs across all samples per class points. The AUSCC is computed using the mean accuracy as the curve.

We use two methods as baselines. A linear SVM with $C = 1.0$ on the raw pixels, and the features produced by the encoder trained in an unsupervised autoencoder, as the feature size $c$ is varied from 4 to 128 features.

Overall results are presented in Table \ref{transfer_results_watertank}. Individual performance with a selected set of layers to produce features are shown in Figure \ref{transfer_watertank_all_models}. We select the best layer for feature extraction based on the highest AUSCC. The best AUSCC is obtained by ResNet20 using the flatten\_5 layer (this is the last layer before the final fully connected layer), with other models closely following with slightly smaller AUSCC, around 1-2\% less.

\begin{table*}[t]
    \centering
    \setlength{\tabcolsep}{6pt}
    \begin{tabular}{lllllllll}
        \toprule
        & & \multicolumn{6}{c}{Test Accuracy}	 & AUSCC\\
        PT Model		 	& Layer 					&  1 Sample & 5 Samples & 10 Samples & 20 Samples & 30 Samples & 50 Samples\\
        \midrule
        ResNet20		 	& flatten\_5 				& $44.6 \pm 7.5\%$ & $76.5 \pm 4.2\%$ & $85.0 \pm 2.8\%$ & $91.9 \pm 1.5\%$ & $94.9 \pm 1.0\%$ & $96.1 \pm 1.1\%$ & \textbf{89.28}\%\\
        & activation\_93			& $39.6 \pm 4.8\%$ & $74.4 \pm 3.0\%$ & $83.5 \pm 2.3\%$ & $89.7 \pm 2.0\%$ & $93.0 \pm 1.6\%$ & $95.6 \pm 0.6\%$ & 87.47\%\\
        & activation\_91			& $33.2 \pm 4.4\%$ & $74.0 \pm 2.6\%$ & $80.9 \pm 2.1\%$ & $87.7 \pm 1.3\%$ & $91.1 \pm 0.8\%$ & $94.1 \pm 1.0\%$ & 85.62\%\\
        \midrule                  
        MobileNet			& conv\_pw\_11\_relu 		&$43.5 \pm 5.0\%$ & $74.5 \pm 2.2\%$ & $84.3 \pm 3.5\%$ & $89.5 \pm 1.0\%$ & $94.1 \pm 0.9\%$ & $95.6 \pm 1.1\%$ & \textbf{88.06}\%\\
        & flatten\_5				& $36.3 \pm 1.7\%$ & $68.8 \pm 2.8\%$ & $77.6 \pm 2.2\%$ & $86.3 \pm 2.6\%$ & $92.2 \pm 1.8\%$ & $94.0 \pm 1.3\%$ & 84.65\%\\
        & conv\_pw\_12\_relu		& $40.3 \pm 7.1\%$ & $70.6 \pm 2.7\%$ & $80.4 \pm 2.7\%$ & $89.2 \pm 1.9\%$ & $92.9 \pm 1.4\%$ & $94.7 \pm 0.7\%$ & 86.46\%\\
        \midrule
        DenseNet121			& conv5\_block15\_0\_relu 	& $41.3 \pm 7.5\%$ & $72.9 \pm 2.8\%$ & $85.8 \pm 2.4\%$ & $92.0 \pm 1.2\%$ & $93.9 \pm 1.0\%$ & $95.8 \pm 1.4\%$ & \textbf{88.76}\%\\
        & conv5\_block16\_0\_relu	& $40.5 \pm 4.6\%$ & $74.0 \pm 4.4\%$ & $83.6 \pm 1.7\%$ & $91.2 \pm 1.9\%$ & $94.7 \pm 1.1\%$ & $96.9 \pm 1.0\%$ & 88.58\%\\
        & avg\_pool					& $33.8 \pm 6.7\%$ & $62.7 \pm 2.2\%$ & $75.2 \pm 1.9\%$ & $83.5 \pm 2.8\%$ & $89.4 \pm 1.7\%$ & $92.5 \pm 0.8\%$ &82.12\%\\
        \midrule
        SqueezeNet			& batch\_norm\_48 	& $39.9 \pm 5.6\%$ & $75.2 \pm 2.9\%$ & $83.0 \pm 1.8\%$ & $90.3 \pm 0.7\%$ & $94.6 \pm 0.6\%$ & $96.7 \pm 0.9\%$ & \textbf{88.48}\%\\
        & batch\_norm\_49 	& $43.6 \pm 3.8\%$ & $72.5 \pm 2.6\%$ & $82.9 \pm 3.0\%$ & $89.9 \pm 1.5\%$ & $93.5 \pm 1.0\%$ & $94.9 \pm 1.0\%$ & 87.54\%\\
        & batch\_norm\_50 	& $29.2 \pm 5.1\%$ & $55.1 \pm 4.5\%$ & $66.0 \pm 2.6\%$ & $73.7 \pm 1.2\%$ & $79.2 \pm 2.3\%$ & $85.2 \pm 1.5\%$ & 72.89\%\\                            
        \midrule
        MiniXception		& add\_19					& $45.5 \pm 6.1\%$ & $75.9 \pm 2.8\%$ & $84.4 \pm 0.8\%$ & $91.4 \pm 1.6\%$ & $93.6 \pm 1.3\%$ & $96.3 \pm 0.9\%$ & \textbf{88.74}\%\\
        & add\_18					& $44.3 \pm 4.7\%$ & $72.5 \pm 3.4\%$ & $81.4 \pm 2.9\%$ & $87.5 \pm 1.1\%$ & $91.8 \pm 1.0\%$ & $95.1 \pm 0.8\%$ & 86.23\%\\
        & conv2d\_35				& $37.5 \pm 5.8\%$ & $66.0 \pm 2.8\%$ & $77.5 \pm 2.0\%$ & $85.5 \pm 2.1\%$ & $89.2 \pm 1.9\%$ & $92.1 \pm 0.9\%$ & 83.04\%\\
        \midrule
        Autoencoder			& enc\_code	(128)			&$46.4 \pm 3.2\%$ & $75.3 \pm 3.2\%$ & $81.0 \pm 2.7\%$ & $86.6 \pm 1.1\%$ & $89.3 \pm 1.9\%$ & $92.3 \pm 1.0\%$  & \textbf{85.11}\%\\	
        & enc\_code (64)			& $45.6 \pm 5.2\%$ & $72.5 \pm 2.6\%$ & $78.6 \pm 1.9\%$ & $83.9 \pm 1.7\%$ & $86.2 \pm 1.2\%$ & $90.1 \pm 1.3\%$ & 82.65\%\\
        & enc\_code (32) 			& $44.3 \pm 8.8\%$ & $68.2 \pm 2.3\%$ & $74.3 \pm 3.1\%$ & $80.4 \pm 2.0\%$ & $82.9 \pm 1.6\%$ & $85.1 \pm 0.6\%$ & 78.60\%\\
        \midrule
        Linear SVM			& NA						& $40.2 \pm 6.1\%$ & $71.5 \pm 4.0\%$ & $81.3 \pm 1.6\%$ & $88.8 \pm 1.0\%$ & $89.6 \pm 1.5\%$ & $92.9 \pm 1.2\%$ & 85.3\%\\
        \toprule
    \end{tabular}
    \caption{Transfer Learning results on the Marine Debris Watertank dataset, as the size of the training set is varied, and comparing multiple feature layers per model. Samples columns indicate how many samples per class are in the training set. AUSCC indicates the area under the sample complexity curve.}
    \label{transfer_results_watertank}
\end{table*}

\newcommand{\plotwitherrorareas}[3]{
    
    \addplot+[mark = none, color=#2] table[x  = spc, y  = accuracy, col sep = semicolon] {#1};
    \addlegendentry{#3}
    
    \addplot [name path=upper, draw=none, forget plot] table[x=spc, y expr=\thisrow{accuracy} + 0.5 * \thisrow{std}, col sep = semicolon] {#1};
    \addplot [name path=lower, draw=none, forget plot] table[x=spc, y expr=\thisrow{accuracy} - 0.5 * \thisrow{std}, col sep = semicolon] {#1};
    \addplot [fill=#2!10, forget plot] fill between[of=upper and lower];   
}

\begin{figure*}[!htb]
    \centering
    \begin{subfigure}{0.32\textwidth}
        \begin{tikzpicture}
            \begin{axis}[height = 0.2 \textheight, width = \textwidth, xlabel={Samples per Class}, ylabel={Test Accuracy}, ymajorgrids=false, xmajorgrids=false, grid style=dashed, legend pos = south east, legend style={font=\scriptsize}, tick label style={font=\scriptsize}, xtick=data]
                
                \plotwitherrorareas{data/tl-watertank-results-fls-turntable-objects-pretrained-densenet121-platform-96x96-layer-avg_pool.csv}{red}{avg\_pool}
                
                \plotwitherrorareas{data/tl-watertank-results-fls-turntable-objects-pretrained-densenet121-platform-96x96-layer-conv5_block16_0_relu.csv}{green}{conv5\_block16\_0\_relu}
                
                \plotwitherrorareas{data/tl-watertank-results-fls-turntable-objects-pretrained-densenet121-platform-96x96-layer-conv5_block15_0_relu.csv}{blue}{conv5\_block15\_0\_relu}
            \end{axis}		            
        \end{tikzpicture}
        \caption{DenseNet121}
    \end{subfigure}
    \begin{subfigure}{0.32\textwidth}
        \begin{tikzpicture}
            \begin{axis}[height = 0.2 \textheight, width = \textwidth, xlabel={Samples per Class}, ylabel={Test Accuracy}, ymajorgrids=false, xmajorgrids=false, grid style=dashed, legend pos = south east, legend style={font=\scriptsize}, tick label style={font=\scriptsize}, xtick=data]
                
                \plotwitherrorareas{data/tl-watertank-results-fls-turntable-objects-pretrained-minixception-platform-96x96-layer-conv2d_35.csv}{red}{conv2d\_35}
                
                \plotwitherrorareas{data/tl-watertank-results-fls-turntable-objects-pretrained-minixception-platform-96x96-layer-add_19.csv}{green}{add\_19}
                
                \plotwitherrorareas{data/tl-watertank-results-fls-turntable-objects-pretrained-minixception-platform-96x96-layer-add_18.csv}{blue}{add\_18}
            \end{axis}		
        \end{tikzpicture}
        \caption{MiniXception}
    \end{subfigure}
    \begin{subfigure}{0.32\textwidth}
        \begin{tikzpicture}
            \begin{axis}[height = 0.2 \textheight, width = \textwidth, xlabel={Samples per Class}, ylabel={Test Accuracy}, ymajorgrids=false, xmajorgrids=false, grid style=dashed, legend pos = south east, legend style={font=\scriptsize}, tick label style={font=\scriptsize}, xtick=data]
                
                \plotwitherrorareas{data/tl-watertank-results-fls-turntable-objects-pretrained-mobilenet-platform-96x96-layer-flatten_5.csv}{red}{flatten\_5}
                
                \plotwitherrorareas{data/tl-watertank-results-fls-turntable-objects-pretrained-mobilenet-platform-96x96-layer-conv_pw_12_relu.csv}{green}{conv\_pw\_12\_relu}
                
                \plotwitherrorareas{data/tl-watertank-results-fls-turntable-objects-pretrained-mobilenet-platform-96x96-layer-conv_pw_11_relu.csv}{blue}{conv\_pw\_11\_relu}
            \end{axis}		
        \end{tikzpicture}
        \caption{MobileNet}
    \end{subfigure}

    \begin{subfigure}{0.32\textwidth}
        \begin{tikzpicture}
            \begin{axis}[height = 0.2 \textheight, width = \textwidth, xlabel={Samples per Class}, ylabel={Test Accuracy}, ymajorgrids=false, xmajorgrids=false, grid style=dashed, legend pos = south east, legend style={font=\scriptsize}, tick label style={font=\scriptsize}, xtick=data]
                
                \plotwitherrorareas{data/tl-watertank-results-fls-turntable-objects-pretrained-resnet20-platform-96x96-layer-flatten_5.csv}{red}{flatten\_5}
                
                \plotwitherrorareas{data/tl-watertank-results-fls-turntable-objects-pretrained-resnet20-platform-96x96-layer-activation_91.csv}{green}{activation\_91}
                
                \plotwitherrorareas{data/tl-watertank-results-fls-turntable-objects-pretrained-resnet20-platform-96x96-layer-activation_93.csv}{blue}{activation\_93}
            \end{axis}		
        \end{tikzpicture}
        \caption{ResNet20}
    \end{subfigure}
    \begin{subfigure}{0.32\textwidth}
        \begin{tikzpicture}
            \begin{axis}[height = 0.2 \textheight, width = \textwidth, xlabel={Samples per Class}, ylabel={Test Accuracy}, ymajorgrids=false, xmajorgrids=false, grid style=dashed, legend pos = south east, legend style={font=\scriptsize}, tick label style={font=\scriptsize}, xtick=data]
                
                \plotwitherrorareas{data/tl-watertank-results-fls-turntable-objects-pretrained-squeezenet-platform-96x96-layer-batch_normalization_48.csv}{red}{batch\_norm\_48}
                
                \plotwitherrorareas{data/tl-watertank-results-fls-turntable-objects-pretrained-squeezenet-platform-96x96-layer-batch_normalization_49.csv}{green}{batch\_norm\_49}
                
                \plotwitherrorareas{data/tl-watertank-results-fls-turntable-objects-pretrained-squeezenet-platform-96x96-layer-batch_normalization_50.csv}{blue}{batch\_norm\_50}
            \end{axis}		
        \end{tikzpicture}
        \caption{SqueezeNet}
    \end{subfigure}
    \begin{subfigure}{0.32\textwidth}
        \begin{tikzpicture}
            \begin{axis}[height = 0.2 \textheight, width = 0.88\textwidth, xlabel={Samples per Class}, ylabel={Test Accuracy}, ymajorgrids=false, xmajorgrids=false, grid style=dashed, legend pos = outer north east, legend style={font=\scriptsize}, tick label style={font=\scriptsize}, xtick=data]
                
                \plotwitherrorareas{data/tl-watertank-results-fls-turntable-objects-pretrained-convencoder-platform-code4-96x96-layer-enc_code.csv}{red}{4}
                
                \plotwitherrorareas{data/tl-watertank-results-fls-turntable-objects-pretrained-convencoder-platform-code8-96x96-layer-enc_code.csv}{green}{8}
                
                \plotwitherrorareas{data/tl-watertank-results-fls-turntable-objects-pretrained-convencoder-platform-code16-96x96-layer-enc_code.csv}{blue}{16}
                
                \plotwitherrorareas{data/tl-watertank-results-fls-turntable-objects-pretrained-convencoder-platform-code32-96x96-layer-enc_code.csv}{black}{32}
                
                \plotwitherrorareas{data/tl-watertank-results-fls-turntable-objects-pretrained-convencoder-platform-code64-96x96-layer-enc_code.csv}{purple}{64}
                
                \plotwitherrorareas{data/tl-watertank-results-fls-turntable-objects-pretrained-convencoder-platform-code128-96x96-layer-enc_code.csv}{brown}{128}
            \end{axis}		
        \end{tikzpicture}
        \caption{Autoencoder}
    \end{subfigure}
    \caption{Comparison of transfer learning on a low-shot setting on the Marine Debris Watertank dataset with multiple models and selected layers for feature extraction. Shaded areas represents one-$\sigma$ confidence intervals.}
    \label{transfer_watertank_all_models}
\end{figure*}
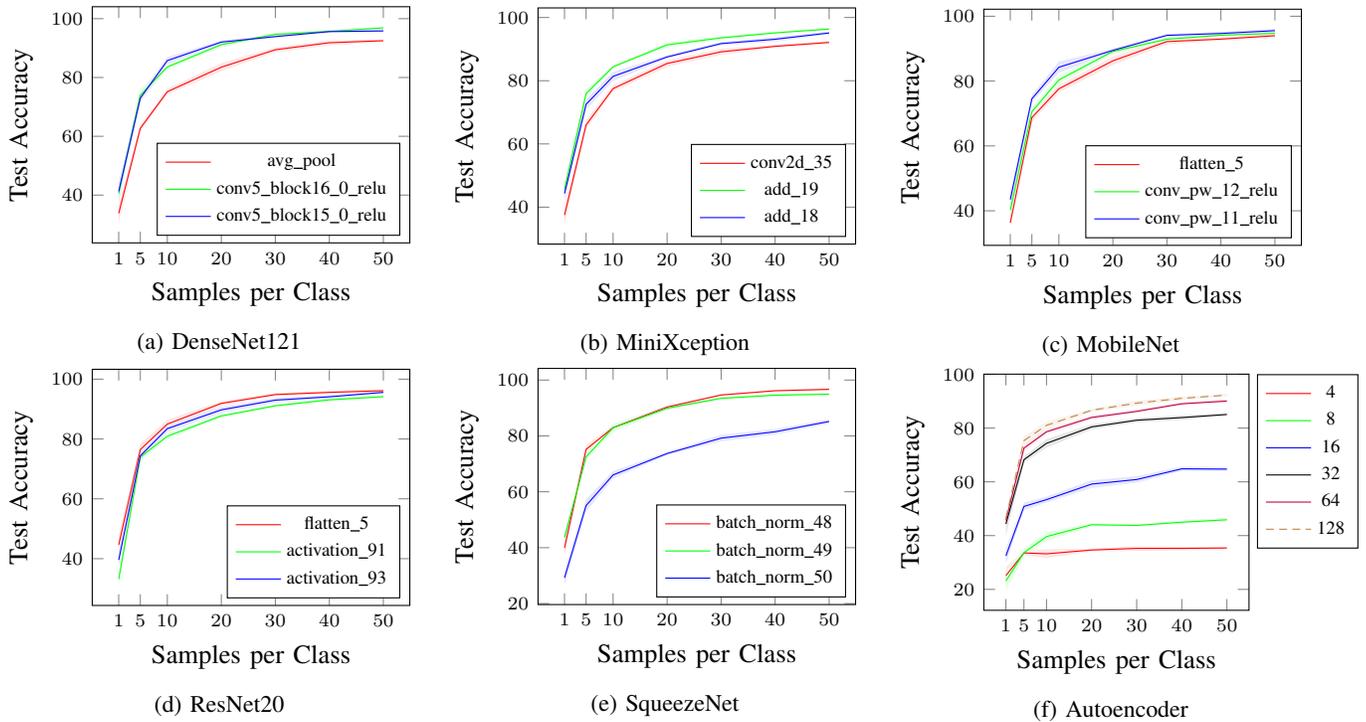

\subsection{Transfer Learning on Gemini 720i Panel-Pipe Dataset}

There is some intersection in the object classes between the watertank and turntable datasets. For a more independent evaluation of the feature quality produced by our pre-trained models, an independent dataset is required. For this purpose we use the Gemini 720i Panel-Pipe dataset, where the object classes are completely independent from the other two datasets, and also this dataset was captured using a different sensor (the Gemini 720i FLS), which can be used to evaluate cross-sensor transferability.

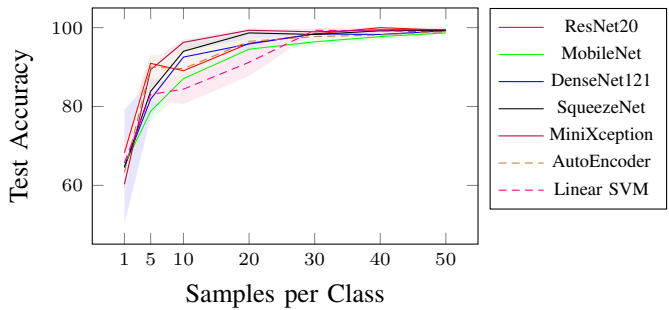
\begin{figure}[htb]
    \centering
    \begin{tikzpicture}
        \begin{axis}[height = 0.2 \textheight, width = 0.37 \textwidth, xlabel={Samples per Class}, ylabel={Test Accuracy}, ymajorgrids=false, xmajorgrids=false, grid style=dashed, legend pos = outer north east, legend style={font=\scriptsize}, tick label style={font=\scriptsize}, xtick=data]
            
            \plotwitherrorareas{data/tl-gemini_720i_panel_pipe-results-fls-turntable-objects-pretrained-resnet20-platform-96x96-layer-flatten_5.csv}{red}{ResNet20}

            \plotwitherrorareas{data/tl-gemini_720i_panel_pipe-results-fls-turntable-objects-pretrained-mobilenet-platform-96x96-layer-conv_pw_11_relu.csv}{green}{MobileNet}
            
            \plotwitherrorareas{data/tl-gemini_720i_panel_pipe-results-fls-turntable-objects-pretrained-densenet121-platform-96x96-layer-conv5_block15_0_relu.csv}{blue}{DenseNet121}
            
            \plotwitherrorareas{data/tl-gemini_720i_panel_pipe-results-fls-turntable-objects-pretrained-squeezenet-platform-96x96-layer-batch_normalization_49.csv}{black}{SqueezeNet}
            
            \plotwitherrorareas{data/tl-gemini_720i_panel_pipe-results-fls-turntable-objects-pretrained-minixception-platform-96x96-layer-add_19.csv}{purple}{MiniXception}
            
            \plotwitherrorareas{data/tl-gemini_720i_panel_pipe-results-fls-turntable-objects-pretrained-convencoder-platform-code128-96x96-layer-enc_code.csv}{brown}{AutoEncoder}
            
            \plotwitherrorareas{data/tl-gemini_720i_panel_pipe-results-linear-svm.csv}{magenta}{Linear SVM}
            
        \end{axis}		
    \end{tikzpicture}
    \caption{Comparison of multiple models on Transfer Learning on the Gemini 720i Panel-Pipe Dataset. Shaded areas represents one-$\sigma$ confidence intervals.}
    \label{transfer_plots_gemini_720i_panel_pipe}
\end{figure}

For this purpose we use the same evaluation protocol as with the watertank dataset, but only using the feature extraction layers that produce the best AUSCC for each model. Results are presented in Table \ref{transfer_results_gemini_720i_panel_pipe} and Figure \ref{transfer_plots_gemini_720i_panel_pipe}.

Overall the best performing model in this  experiment is MiniXception, with AUSCC of 96.31\%, closely followed by SqueezeNet. There is a large variation across different number of samples in the training set, as ResNet20 performs better in the low-shot setting at one samples per class, but slightly under-performing for higher samples per class (like 10 and 20 samples).

\begin{table*}[!t]    
    \centering
    \setlength{\tabcolsep}{5pt}
    \begin{tabular}{lllllllll}
        \toprule
        & & \multicolumn{6}{c}{Test Accuracy}	 & AUSCC\\
        PT Model 	& Layer 					&  1 Sample & 5 Samples & 10 Samples & 20 Samples & 30 Samples & 50 Samples\\
        \midrule
        ResNet20		 	& flatten\_5 				& $\mathbf{68.3 \pm 11.4}\%$ & $90.9 \pm 4.4\%$ & $89.1 \pm 5.6\%$ & $96.0 \pm 4.1\%$ & $98.4 \pm 2.9\%$ & $99.3 \pm 1.3\%$ & 95.00\%\\
        MobileNet			& conv\_pw\_11\_relu 		& $65.1 \pm 9.7\%$ & $78.8 \pm 4.7\%$ & $87.1 \pm 6.4\%$ & $94.5 \pm 2.6\%$ & $96.4 \pm 1.8\%$ & $98.7 \pm 0.4\%$ & 92.20\%\\
        DenseNet121			& conv5\_block15\_0\_relu 	& $64.7 \pm 29.1\%$ & $81.9 \pm 9.1\%$ & $92.5 \pm 6.1\%$ & $95.9 \pm 3.6\%$ & $98.4 \pm 1.1\%$ & $99.2 \pm 0.7\%$ & 94.14\%\\
        SqueezeNet			& batch\_norm\_48 	& $60.9 \pm 7.6\%$ & $84.4 \pm 9.7\%$ & $\mathbf{95.9 \pm 4.5}\%$ & $96.1 \pm 4.6\%$ & $\mathbf{99.5 \pm 0.8}\%$ & $\mathbf{100.0 \pm 0.0}\%$ & 95.25\%\\
        MiniXception		& add\_19					& $60.3 \pm 5.7\%$ & $89.3 \pm 4.3\%$ & $\mathbf{96.3 \pm 1.4}\%$ & $\mathbf{99.3 \pm 1.0}\%$ & $\mathbf{98.9 \pm 1.3}\%$ & $\mathbf{99.5 \pm 1.1}\%$  & \textbf{96.31}\%\\
        \midrule
        Autoencoder			& enc\_code					& $63.3 \pm 7.3\%$ & $90.0 \pm 5.6\%$ & $89.6 \pm 8.8\%$ & $96.5 \pm 1.4\%$ & $97.7 \pm 1.2\%$ & $99.1 \pm 1.0\%$ & 94.35\%\\	
        \midrule
        Linear SVM			& NA						& $65.7 \pm 8.8\%$ & $83.1\% \pm 3.3$ & $84.4 \pm 7.5\%$ & $91.2 \pm 6.9\%$ & $99.3 \pm 1.0\%$ & $98.9 \pm 1.3\%$ & 92.45\%\\
        \toprule
    \end{tabular}
    \caption{Transfer Learning results on the Gemini 720i Panel-Pipe dataset, using the best feature layers, as the size of the training set is varied. Samples columns indicate how many samples per class are in the training set. AUSCC indicates the area under the sample complexity curve.}
    \label{transfer_results_gemini_720i_panel_pipe}
\end{table*}

\subsection{Discussion and Analysis}

In both experiments we see an increase in the variation of accuracy with less samples, with one sample per class having the highest standard deviation, indicating that while good performance can be achieved, it depends on which exact samples are in the training set. More samples stabilize training.

The SVM baselines trained on raw pixel data are overall the weakest in performance, showing the importance of learning appropriate features specific to sonar images.

We believe that our results show that transfer learning using these pre-trained models can help reduce sample complexity, particularly in low-shot settings that are common in underwater perception.


Of the architectures we have selected, there is no particular architecture that outperforms the rest in all cases. This is known as the free lunch theorem in the ML community, where in average no ML model outperforms the rest. This motivates the selection and use of multiple models in this work, as some models will work better in some situations, while other architectures might perform adequately in other situations or modalities. For example, ResNet20 works quite well in the low-shot setting in the Gemini 720i Panel-Pipe dataset, but when more samples are available (from around 10 samples per class), it is outperformed by SqueezeNet and MiniXception, while ResNet20 overall has the best performance in the Marine Debris Watertank dataset.

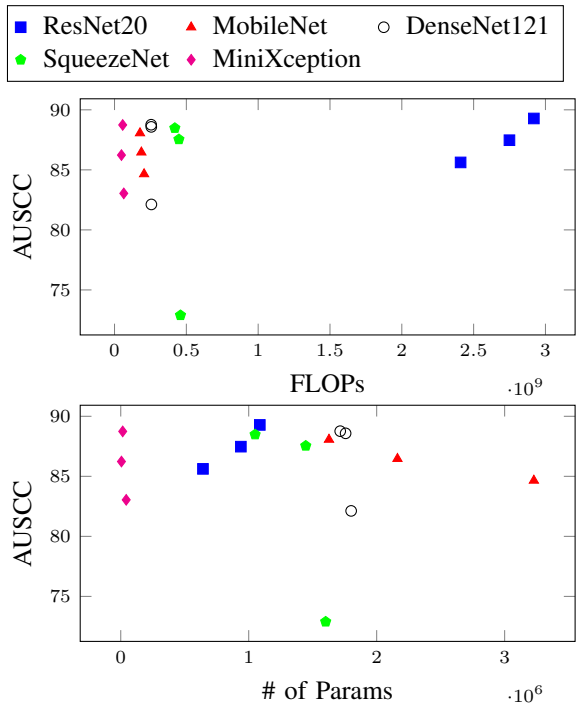
\begin{figure}
    \centering
    \begin{tikzpicture}
        \begin{customlegend}[legend columns = 3,legend style = {column sep=1ex}, legend cell align = left,
            legend entries={ResNet20, MobileNet, DenseNet121, SqueezeNet, MiniXception}]
            \addlegendimage{mark=square*,blue, only marks}
            \addlegendimage{mark=triangle*,red, only marks}
            \addlegendimage{mark=o,black, only marks}
            \addlegendimage{mark=pentagon*,green, only marks}
            \addlegendimage{mark=diamond*,magenta, only marks}
        \end{customlegend}
    \end{tikzpicture}
    \vspace*{0.5em}
    
    \begin{tikzpicture}
        \begin{axis}[height = 0.2 \textheight, width = 0.45 \textwidth, xlabel={FLOPs}, ylabel={AUSCC}, ymajorgrids=false, xmajorgrids=false, grid style=dashed, legend pos = outer north east, legend style={font=\scriptsize}, tick label style={font=\scriptsize}]
            \addplot[scatter, only marks, scatter src=explicit symbolic,
                     scatter/classes={resnet20={mark=square*,blue},mobilenet={mark=triangle*,red},densenet121={mark=o,draw=black}, squeezenet={mark=pentagon*,green}, minixception={mark=diamond*,magenta}}
            ] table[meta=model, x  = flops, y  = auc] {
                auc		flops		model
                89.28	2919350016	resnet20
                87.47	2749332096	resnet20
                85.62	2409296256	resnet20
                88.06	177354720	mobilenet
                84.65	205915176	mobilenet
                86.46	186874884	mobilenet
                88.76	255148146	densenet121
                88.58	255867030	densenet121
                82.12	256604346	densenet121
                88.48	419853024	squeezenet
                87.54	448210944	squeezenet
                72.89	459271080	squeezenet
                88.74	56898720	minixception
                86.23	48233088	minixception
                83.04	64474848	minixception                
            };
        \end{axis}
    \end{tikzpicture}

    \begin{tikzpicture}
        \begin{axis}[height = 0.2 \textheight, width = 0.45 \textwidth, xlabel={\# of Params}, ylabel={AUSCC}, ymajorgrids=false, xmajorgrids=false, grid style=dashed, legend pos = outer north east, legend style={font=\scriptsize}, tick label style={font=\scriptsize}]
            \addplot[scatter, only marks, scatter src=explicit symbolic,
            scatter/classes={resnet20={mark=square*,blue},mobilenet={mark=triangle*,red},densenet121={mark=o,draw=black}, squeezenet={mark=pentagon*,green}, minixception={mark=diamond*,magenta}}
            ] table[meta=model, x  = params, y  = auc] {
                auc		params	model
                89.28	1086656	resnet20
                87.47	938560	resnet20
                85.62	642368	resnet20
                88.06	1627264	mobilenet
                84.65	3228288	mobilenet
                86.46	2162304	mobilenet
                88.76	1714976	densenet121
                88.58	1757152	densenet121
                82.12	1800416	densenet121
                88.48	1049776	squeezenet
                87.54	1445680	squeezenet
                72.89	1601340	squeezenet
                88.74	15256	minixception
                86.23	5432	minixception
                83.04	42584	minixception
                
            };
        \end{axis}
    \end{tikzpicture}
    \caption{Comparison of all architectures and selected layers on the Watertank dataset, in terms of AUSCC versus FLOPs and number of parameters in the model.}
    \label{overall_comparison_watertank}
\end{figure}

\begin{figure}
    \centering
    \begin{tikzpicture}
        \begin{customlegend}[legend columns = 3,legend style = {column sep=1ex}, legend cell align = left,
            legend entries={ResNet20, MobileNet, DenseNet121, SqueezeNet, MiniXception}]
            \addlegendimage{mark=square*,blue, only marks}
            \addlegendimage{mark=triangle*,red, only marks}
            \addlegendimage{mark=o,black, only marks}
            \addlegendimage{mark=pentagon*,green, only marks}
            \addlegendimage{mark=diamond*,magenta, only marks}
        \end{customlegend}
    \end{tikzpicture}
    \vspace*{0.5em}
    
    \begin{tikzpicture}
        \begin{axis}[height = 0.2 \textheight, width = 0.45 \textwidth, xlabel={FLOPs}, ylabel={AUSCC}, ymajorgrids=false, xmajorgrids=false, grid style=dashed, legend pos = outer north east, legend style={font=\scriptsize}, tick label style={font=\scriptsize}]
            \addplot[scatter, only marks, scatter src=explicit symbolic,
            scatter/classes={resnet20={mark=square*,blue},mobilenet={mark=triangle*,red},densenet121={mark=o,draw=black}, squeezenet={mark=pentagon*,green}, minixception={mark=diamond*,magenta}}
            ] table[meta=model, x  = flops, y  = auc] {
                auc		flops		model
                95.00	2919350016	resnet20
                92.20	177354720	mobilenet
                94.14	255148146	densenet121
                95.25	419853024	squeezenet
                96.31	56898720	minixception
            };
        \end{axis}
    \end{tikzpicture}
    
    \begin{tikzpicture}
        \begin{axis}[height = 0.2 \textheight, width = 0.45 \textwidth, xlabel={\# of Params}, ylabel={AUSCC}, ymajorgrids=false, xmajorgrids=false, grid style=dashed, legend pos = outer north east, legend style={font=\scriptsize}, tick label style={font=\scriptsize}]
            \addplot[scatter, only marks, scatter src=explicit symbolic,
            scatter/classes={resnet20={mark=square*,blue},mobilenet={mark=triangle*,red},densenet121={mark=o,draw=black}, squeezenet={mark=pentagon*,green}, minixception={mark=diamond*,magenta}}
            ] table[meta=model, x  = params, y  = auc] {
                auc		params	model
                95.00	1086656	resnet20
                92.20	1627264	mobilenet
                94.14	1714976	densenet121
                95.25	1049776	squeezenet
                96.31	15256	minixception                
            };
        \end{axis}
    \end{tikzpicture}
    \caption{Comparison of all architectures with their best layer on the Gemini 720i Panel-Pipe dataset, in terms of AUSCC versus FLOPs and number of parameters in the model.}
    \label{overall_comparison_gemini}
\end{figure}
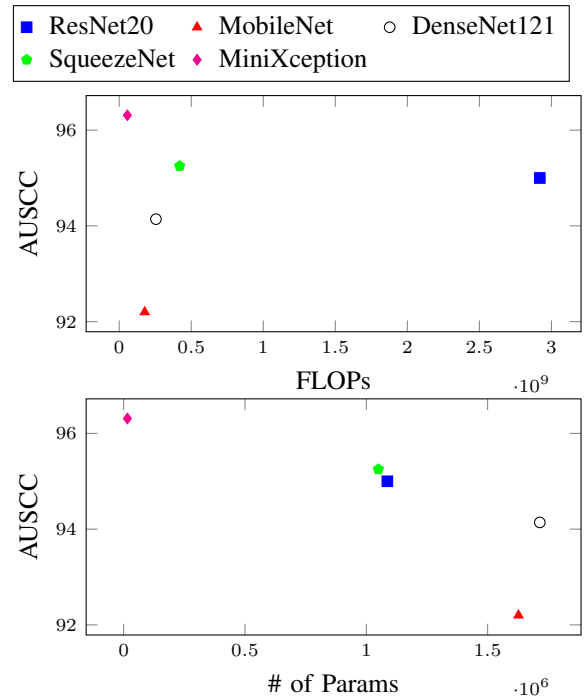

For the future user of these models, it is important to always evaluate multiple models and choose the one that fits their needs in terms of task performance, computation time, etc. Figures \ref{overall_comparison_watertank} and \ref{overall_comparison_gemini} can help make this decision.

\section{Conclusions and Future Work}

In this paper we have presented our concept of pre-trained convolutional neural networks for sonar images, as to fill the missing gap of pre-trained models specifically designed and tuned for sonar images.  We trained our models on the turntable dataset, and evaluated our these models in the task of low-shot sonar image classification on the watertank (some object classes are common between turntable and watertank datasets) and panel-pipe datasets (no classes in common), showing that the learned features are useful for classification with low number of samples per class, outperforming the autoencoder and linear SVM baselines.

We expect that our pre-trained models will be used in autonomous underwater vehicles, as they are tuned for maximum classification and computational performance. This allows the designer to select the most appropriate architecture and input image size, and our paper provides information about the trade-offs between task and computational performance.

All pre-trained models in this paper are publicly available at \url{https://github.com/mvaldenegro/pretrained-models-sonar-images/} as Keras\footnote{Trained using Keras 2.3.1 on TensorFlow 1.14.0} HDF5 model files, which can be loaded in other frameworks and should serve as a basis for future research.

\section*{Acknowledgements}

The authors would like to thank Leonard McLean from the Ocean Systems Lab, Heriot-Watt University, for his help in capturing data used
in this paper.

\bibliographystyle{ieeetr} 
\bibliography{biblio.bib}

\end{document}